%% file: main.tex
\documentclass{article}
\input{iclr-preamble}
\iclrfinalcopy
\hypersetup{
  pdftitle={RoomTokens: Masked Diffusion over Structured Tokens for Indoor Scene Synthesis},
  pdfauthor={Xiaohao Sun and Angel X. Chang}}

\input{figs/fig-teaser}
\begin{document}

\title{\Large RoomTokens: Masked Diffusion over\\Structured Tokens for Indoor Scene Synthesis}

\author{Xiaohao Sun\textsuperscript{1} \qquad Angel X. Chang\textsuperscript{1,2}\\
  \textsuperscript{1}Simon Fraser University\\
  \textsuperscript{2}Alberta Machine Intelligence Institute (Amii)\\[2pt]
  \href{https://3dlg-hcvc.github.io/RoomToken/}{\textcolor[HTML]{A52C62}{\normalfont\small https://3dlg-hcvc.github.io/RoomToken/}}}

\maketitle
\pagestyle{plain}

\input{sections/00-abstract}

\input{sections/01-introduction}
\input{sections/02-related-work}
\input{sections/03-method}
\input{sections/04-experiments}

\input{sections/05-conclusion}

\input{sections/acknowledgments}

\section*{AI use statement}
We used AI tools for the following tasks:
\begin{itemize}
    \item Help develop conceptual frameworks: Coding agents helped brainstorm alternatives for the scene representation and related method design. The authors selected and refined the final approach.
    \item Design or provide feedback on research methodology or experiments: Coding agents provided feedback on the evaluation protocol, metrics, and ablation experiments.
    \item Implement methods: Coding agents assisted with implementing and debugging the model, training, scene assembly, and evaluation code.
    \item Support qualitative and thematic data analysis: Coding agents helped examine experimental results and ablations, identify inconsistencies, and interpret patterns in the reported metrics.
\end{itemize}

Additional disclosures include the use of AI for brainstorming, creating and editing software code, searching for information and relevant literature, and drafting and editing manuscript text to improve clarity. AI also helped clarify the written and mathematical descriptions of our training objectives and evaluation metrics.

We have reviewed and validated all AI-assisted work and take responsibility for the final content of the work, including text, claims, code, and other artifacts produced with the aid of AI.

\bibliographystyle{plainnat}
\bibliography{main}

\appendix
\clearpage
\newpage
\input{sections/06-appendix}

\end{document}

%% file: figs/fig-teaser.tex
\newcommand{\figteasercontent}{
\begin{center}
  \includegraphics[trim={0 4pt 0 0},clip,width=\textwidth]{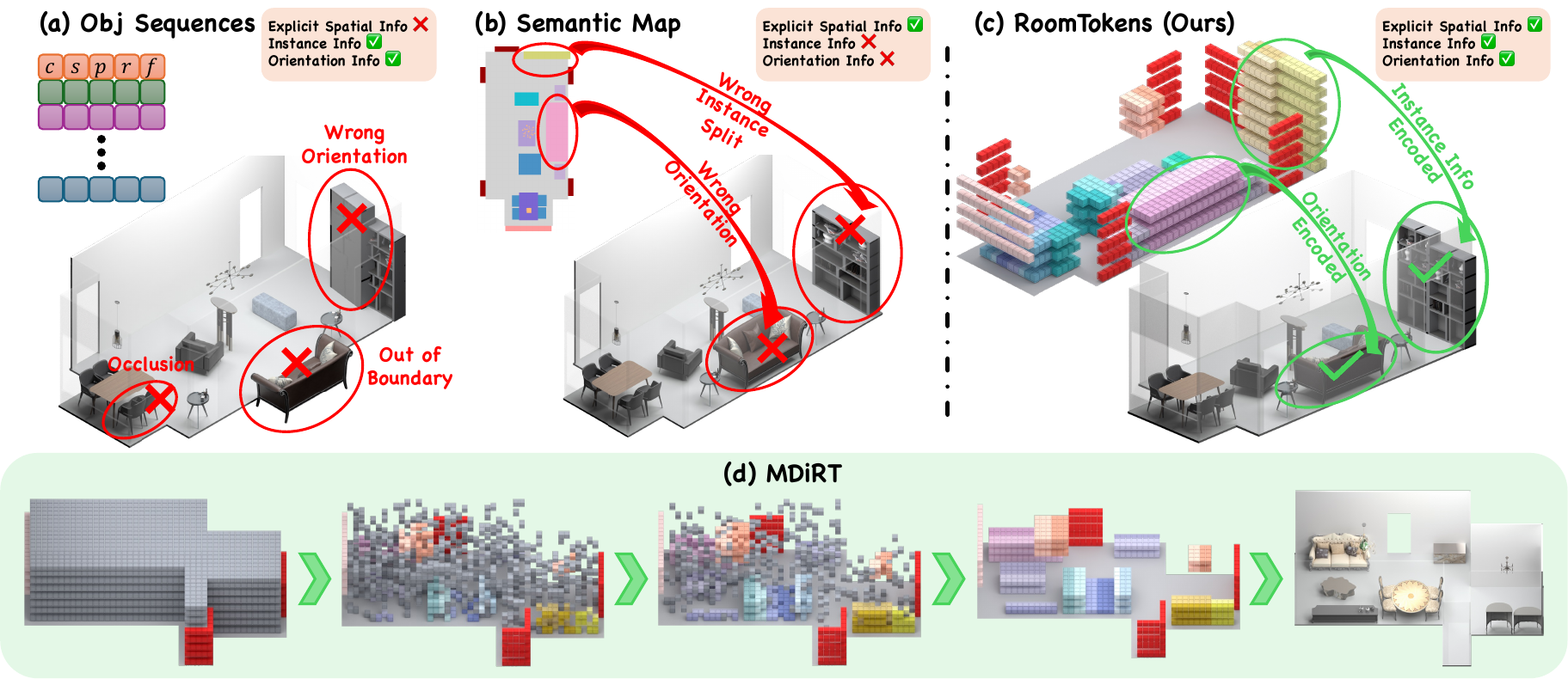}
  \captionof{figure}{\textit{Top.}
  Comparison of scene representations for indoor scene generation. Object sequences (a) encode object attributes but lack explicit spatial relationship information. Semantic maps (b) make spatial structure explicit, but lack vertical, instance, and orientation information. We propose \repname (c) to address these missing factors by encoding semantic occupancy, instance identity, and facing cues in a sliced spatial grid.
  \textit{Bottom.} We introduce \methodname{} (d), a masked discrete diffusion model over \repname, which denoises missing token fields while keeping architectural tokens fixed so generated scenes respect room boundaries.
  \vspace{-2em}
  }
  \label{fig:teaser}
\end{center}
}

\newcommand{\figteaser}{%
\begin{figure}
\vspace{-1em}
\figteasercontent
\end{figure}
}

%% file: sections/00-abstract.tex
\begin{abstract}
Generating 3D indoor scenes requires a representation that captures object semantics, instance identity, height structure, and orientation while respecting room geometry.
Methods that model scenes as sequences of object vectors learn spatial constraints implicitly, leaving collisions and boundary violations to losses or post-processing; semantic-map methods make occupancy explicit, but collapse instance and directional structure. We introduce \repname{}, a sliced structured-token representation in which each token stores semantic class, instance-color identity, and facing field. This representation allows a scene to be generated as a discrete grid while preserving cues for instance grouping, orientation estimation, and 3D extent prediction. To learn the RoomToken distribution, we propose \methodname{} (Masked Diffusion over RoomTokens), a masked discrete diffusion model that denoises semantic/instance-color/facing channels jointly with a shared cross-slice backbone. Architectural tokens are clamped as explicit floor-plan constraints, so the model samples furniture layouts that respect room geometry while retaining instance and object-attribute structure.
Experiments on 3D-FRONT show that RoomTokens with \methodname{} improve layout fidelity and physical consistency over sequence/vector- and map-based methods, suggesting structured spatial tokens as a practical choice for discrete 3D scene synthesis.
\end{abstract}

%% file: sections/01-introduction.tex
\vspace{-0.5em}
\section{Introduction}
\vspace{-0.5em}
Automatic generation of 3D scenes is increasingly important, as it allows for scaling up of training data for 3D scene understanding and embodied AI, and provides content for applications such as augmented reality, gaming, and entertainment.
Compositional 3D indoor scene generation~\citep{tam2026survey} formulates the problem as an object arrangement problem, where we must decide on object categories, individual object identities, placement, and orientation while respecting the geometry of the room.
To ensure that generated scenes are plausible, it is important to model the spatial structure and object-level semantics together.

Most existing methods represent scenes as sets or sequences of object vectors~\citep{paschalidou2021atiss,tang2024diffuscene,hu2026mixed,lin2024instructscene}, where each object vector encodes the object category and bounding box of the object.
While this representation is compact,
it does not make occupancy constraints explicit and cannot ensure that two generated boxes do not occupy the same space.
As a result, generated scenes may contain inter-object collisions, wall penetrations, and inadequate clearance for accessing objects and room regions. Prior systems therefore add auxiliary mechanisms, such as intersection penalties in the training objective~\citep{tang2024diffuscene}, rejection-style resampling of invalid insertions~\citep{ritchie2019fast}, or optimization-based layout refinement after an initial prediction~\citep{luo2020end}.
Another line of work~\citep{feng2023layoutgpt,yang2025sceneweaver,xia2026sage} leverages the broad knowledge encoded in LLMs/VLMs for more extensive semantic priors, but their coordinate outputs still require geometric cleanup and potentially multiple passes with a visual critic to ensure a physically plausible scene.

Approaches that generate scenes based on semantic-maps~\citep{yang2021indoor,sun2026semlayoutdiff} capture the spatial arrangements explicitly by assigning semantic labels to cells on a dense grid.
A recent work, SemLayoutDiff~\citep{sun2026semlayoutdiff} demonstrated improved floor-plan-conditioned generation with fewer collisions and out-of-boundary objects.
However, a single top-down semantic map fails to capture important information needed for 3D scene generation.
It collapses vertical information, merges adjacent objects of the same category under connected-component decoding, and provides no directional cue for estimating object orientation.
In this work, we tackle the challenge of how to introduce instance-level and vertical information into these spatial maps.
\figteaser

We introduce \textbf{\repname}, a sliced structured-token representation for indoor scenes. We represent a room as a stack of height-aware grids, with each grid cell storing a RoomToken with three fields that capture the semantic class, the instance identity, and facing cues.
Encoding the instance identity is challenging as the assignment of a particular ID to an object is arbitrary (i.e., can be permuted or re-assigned), and the number of objects in a scene is not fixed.
As our aim is to distinguish spatially adjacent object instances on a planar map (vs. providing an instance ID), we can use four `colors' (following the four-color theorem~\citep{robertson1997four}) to identify different instances.
For orientation, the facing field encodes whether a furniture token occupies the body or front half of an object; non-furniture tokens use none. We find that these two furniture labels are sufficient for decoding orientation.
Together, these properties provide sufficient information for a model to generate a discrete spatial layout while preserving cues for instance grouping, orientation estimation, and 3D extent prediction.

To learn the RoomToken distribution, we propose \textbf{\methodname} (Masked Diffusion over RoomTokens), a single-stage masked discrete diffusion model for indoor scene generation. Each token property is embedded as a semantic, instance-color, or facing channel.
The channels are processed by a shared cross-slice UNet and decoded jointly. The denoising process maps masked RoomTokens to structured 3D scene layouts while conditioning on and clamping the architectural tokens from the floor plan (\cref{fig:teaser} (bottom)), so furniture tokens are generated within the valid interior region defined by the architectural condition. Instead of predicting semantics and then instance-level attributes as in prior work~\citep{sun2026semlayoutdiff}, this multichannel parameterization learns semantic, instance-color, and facing dependencies through shared features in a single reverse process.
Experiments on 3D-FRONT~\citep{fu20213dfront} show that \methodname{} generates more plausible layouts compared with \semdiff{}~\citep{sun2026semlayoutdiff}. Generated layouts better match reference rooms (FID 50.14 vs.~71.18) and capture more plausible object relations (Pair-Rel 12.09 vs.~20.65). They also contain fewer collisions (COL 12.64 vs.~16.24) and leave doors and windows more unobstructed (OC 82.31 vs.~56.02).

In summary, our contributions are:
1)    \textbf{A sliced structured RoomToken representation} that stores semantic class, graph-color instance identity, and facing information in each height-aware grid token.
2)    \textbf{A single-stage masked discrete diffusion model, \methodname{}, for RoomTokens} that clamps the architectural condition throughout denoising so furniture is generated within the valid room interior, while jointly modeling semantic/instance-color/facing channels with a shared cross-slice backbone.
3)    \textbf{An empirical analysis} showing that explicit instance-color and facing fields improve instance recovery and orientation estimation and provide for physically plausible scene layouts.

%% file: sections/02-related-work.tex
\section{Related Work}
\label{sec:related}
\mypara{Object-vector and relational scene synthesis.}
Many indoor scene synthesis methods represent a room as a set or sequence of objects with attributes specifying the semantic class, position, and orientation of objects. %
Early neural and graph-based approaches learn spatial priors over object arrangements~\citep{wang2018deep,ritchie2019fast,li2019grains,wang2019planit,zhang2020deep}, while recent autoregressive and diffusion methods predict the object attributes directly~\citep{wang2021sceneformer,paschalidou2021atiss,sun2024forest2seq,tang2024diffuscene,hu2026mixed,maillard2024debara,sun2025reltriple}.
Other recent work leverages scene graphs, relational triples, or mixed graph diffusion for improved modeling of object layouts~\citep{lin2024instructscene,zhai2023commonscenes,zhai2024echoscene,bai2025freescene,ruiz2025geoscenegraph},
and physical or human-aware constraint objectives and optimization~\citep{yi2023mime,yang2024physcene,sun2024haisor,su2025chord,lin2026pat3d}.
These methods differ in conditioning and priors, but their generated layout is ultimately a set of object bounding boxes.
As a result, generated scenes may contain object collisions and boundary overlaps, and additional post-processing is required as the representation does not natively prevent these.
\methodname{} instead generates a discrete occupancy grid, where each cell has a single semantic assignment and grid-level overlap is therefore explicit. By modeling object occupancy explicitly across height-aware slices, the representation helps reduce inter-object collisions.

\mypara{LLM and agent-based scene generation.}
By leveraging LLMs and VLMs, it is possible to condition on language input and access the common sense knowledge of how objects are positioned, stored in these pretrained models.
LayoutGPT~\citep{feng2023layoutgpt} predicts layouts with language models, while Holodeck~\citep{yang2024holodeck}, AnyHome~\citep{wen2024anyhome}, SceneTeller~\citep{ocal2024sceneteller}, and open-universe systems~\citep{aguina2024open} connect text prompts with object retrieval, constraint solving, or scene assembly.
More recent LLM/VLM and agentic methods use program synthesis, visual-language feedback, or multi-agent refinement for broader controllability~\citep{hu2024scenecraft,ccelen2024design,sun2025layoutvlm,yang2025sceneweaver,pfaff2026scenesmith,gao2025discolayout}.
These systems provide flexible interfaces, but coordinate- or program-level outputs still require geometric cleanup for dense indoor layouts.
Our focus is complementary: we learn the layout distribution from annotated 3D-FRONT rooms and make the representation itself carry local occupancy, instance, and orientation structure.

\mypara{Discrete spatial representations for scene synthesis.}
Discrete spatial representations encode the scene by assigning labels directly to locations on a 2D or 3D lattice, making occupancy constraints local and explicit.
\citet{yang2021indoor} use voxel labels, \citet{lee2023diffusion} model scene-scale categorical 3D data, and SemLayoutDiff~\citep{sun2026semlayoutdiff} applies multinomial discrete diffusion to a top-down semantic map.
This family is closest to ours because it moves generation from unordered boxes toward explicit spatial fields.
However, semantic-only spatial grids do not distinguish touching same-class instances, top-down maps collapse vertical structure, and a single class label per cell provides no direct orientation cue.
Our representation keeps the grid advantage but upgrades each sliced cell into a structured RoomToken carrying semantic, instance-color, and facing fields.
Graph coloring is used to assign instance colors during preprocessing, with the four-color theorem~\citep{appel1977every,robertson1997four} providing the minimum number of colors needed for separating adjacent same-class instances. Similar coloring ideas have been studied for discriminative instance segmentation~\citep{kulikov2018instance,zhangfour}, but remain largely unexplored as compact instance channels for generative modelling.

\mypara{Discrete diffusion and masked token modeling.}
Masked discrete diffusion corrupts categorical data by replacing tokens with absorbing mask states and trains a denoiser to reverse that process.
Multinomial and masked variants~\citep{hoogeboom2021argmax,shi2024simplified,lou2024sedd,sahoo2024mdlm,nie2025llada} have become effective for discrete sequences and token grids.
Layout-specific diffusion models also use discrete or mixed discrete-continuous variables for structured 2D layouts and indoor scene attributes~\citep{chai2023layoutdm,inoue2023layoutdm,levi2023dlt,hui2023unifying,choi2026scenat}, but these variables are often organized as layout tokens or object attributes rather than as per-cell scene occupancy fields.
We extend masked discrete diffusion from a single semantic map or object-attribute tokenization to structured RoomTokens.
\methodname{} uses a multichannel parameterization for training, where semantic, instance-color, and facing values are denoised together under validity constraints and floor-plan conditioning.

%% file: sections/03-method.tex
\section{Method}
\label{sec:method}
\input{figs/fig-representation}
\input{figs/fig-method}

\vspace{-0.5em}
Given a room type $r$ and an input architectural mask (arch mask), \methodname{} generates a complete object-level 3D layout (\cref{fig:method}). With our RoomToken representation, slices at different heights store semantic occupancy with instance and facing cues, allowing objects to be recovered from the grid tokens (\cref{sec:representation}).
We use masked diffusion to generate interior RoomTokens while keeping the architectural tokens fixed (\cref{sec:diffusion}). At each reverse step, a multichannel denoising network $g_\theta$ predicts the masked fields across height slices (\cref{sec:denoiser}). Finally, during scene assembly, the token grid is converted into object records and matching assets are retrieved for the final scene (\cref{sec:inference}).

\vspace{-0.5em}
\subsection{Sliced RoomTokens}
\label{sec:representation}

To recover object instances and poses from a discrete occupancy map, we encode each room as a stack of sliced RoomTokens.
We discretize the floor plane into an $H\times W$ grid and use $S$ selected height slices $\mathcal{Z}=\{\zeta_1,\ldots,\zeta_S\}$ to capture the vertical extent (\cref{fig:representation}). We choose these heights by greedy coverage over furniture height intervals in the training set: each object covers a candidate height $h$ if its vertical interval intersects $[h-\tau,h+\tau]$, with $\tau{=}0.10$m matching the rasterization tolerance. Starting from fixed floor and ceiling anchors, we repeatedly add the candidate height that covers the most previously uncovered objects. We provide the specific grid size, slice heights, and other parameters in \cref{sec:exp_setup}.

We index a token by $i=(u,v,s)$, where $u\in\{1,\ldots,H\}$, $v\in\{1,\ldots,W\}$, and $s\in\{1,\ldots,S\}$. Each grid location stores a token $\mathbf{x}_i=(y_i,c_i,f_i)$ with semantic class $y_i$, instance-color label $c_i$, and facing label $f_i$. Semantic labels use $K_{\mathrm{sem}}$ classes. The instance-color label is either one of $C$ instance colors or the none value $\varnothing_c$; the facing label is body, front, or the none value $\varnothing_f$.
The three values must agree with the semantic class.
Let $Y_{\mathrm{arch}}$ denote architecture/background classes and $Y_{\mathrm{furn}}$ denote furniture classes. A RoomToken $\mathbf{x}_i=(y_i,c_i,f_i)$ is valid only if
\begin{equation}
  \mathrm{valid}(y_i,c_i,f_i) \Longleftrightarrow
  \begin{cases}
  c_i=\varnothing_c \text{ and } f_i=\varnothing_f, & y_i\in Y_{\mathrm{arch}}, \\
  c_i\in\{0,\ldots,C-1\} \text{ and } f_i\in\{\mathrm{body},\mathrm{front}\}, & y_i\in Y_{\mathrm{furn}}.
  \end{cases}
  \label{eq:valid_support}
\end{equation}
This rule ties instance and facing cues to semantic occupancy: architecture/background tokens carry no attributes, while furniture tokens carry both instance color and facing labels.

\vspace{-1em}
\subsection{Masked Diffusion over RoomTokens}
\label{sec:diffusion}

Because the arch mask already specifies the room boundary, the diffusion process should generate only the unknown interior while leaving the observed architecture unchanged.
To accomplish this, we use masked diffusion~\citep{shi2024simplified}.
In contrast to regular diffusion models that slowly add noise to the input, masked diffusion gradually replaces and hides the interior RoomToken labels with a mask token during the forward process.  See \cref{fig:teaser} bottom for the reverse process.

\mypara{Forward process.} We formulate the forward masking process from a clean RoomToken cell $\mathbf{x}$ to the partially masked $\mathbf{z}_t$ at time step $t$ as repeatedly sampling whether to keep the original token or replace it with a mask token.
For our three channel RoomToken $\mathbf{x}_i = (y_i, c_i, f_i)$, we will use three mask tokens $[\mathrm{M}]_y,[\mathrm{M}]_c,[\mathrm{M}]_f$.
We draw independent keep indicators $m_i$ for the semantic label, $n_i$ for the instance-color and facing labels from a Bernoulli distribution with keep probability $\alpha(t)$:  $m_i,n_i\sim\mathrm{Bernoulli}(\alpha(t))$.
Both instance-color and facing labels are revealed only when the semantic label is visible. The indicator $a_i=1$ marks fixed architectural cells, while $a_i=0$ marks cells to generate.
For $\rho\in\{c,f\}$, let $x_i^c=c_i$ and $x_i^f=f_i$.
The forward mask process
$q_t(\mathbf{z}_t\mid\mathbf{x},\mathbf{a})$ is
\begin{equation}
  \begin{aligned}
  z^y_{t,i} &=
  \begin{cases}
    y_i, & \text{if } a_i=1 \text{ or } m_i=1, \\
    [\mathrm{M}]_y, & \text{otherwise,}
  \end{cases}
  &&
  z^\rho_{t,i} &=
  \begin{cases}
    \varnothing_\rho, & \text{if } a_i=1, \\
    x_i^\rho, & \text{if } a_i=0 \text{ and } m_i=n_i=1, \\
    [\mathrm{M}]_\rho, & \text{otherwise.}
  \end{cases}
  \end{aligned}
  \label{eq:forward_process}
\end{equation}
\Cref{eq:forward_process} masks both attributes whenever an unfixed cell's semantic label is masked, regardless of its ground-truth category. Sampling initializes all unfixed cells with all three fields masked, while the supplied architecture remains fixed.

\mypara{Reverse process.} The reverse process fills masked tokens by stepping from time $t$ to an earlier denoising time $s<t$ under the same constraints. We condition on $\kappa=(r,\mathbf{a})$, where $r$ is the room type and $\mathbf{a}$ is the arch-mask indicator. Given the partially masked grid $\mathbf{z}_t$, time $t$, and condition $\kappa$, the network $g_\theta(\mathbf{z}_t,t,\kappa)$ predicts label probabilities for the semantic, instance-color, and facing fields. We use these probabilities to sample the proposal $\hat{\mathbf{z}}_s$. We then restore fixed arch-mask tokens and project generated tokens to the valid RoomToken support:
\begin{equation}
  \mathbf{z}_{s,i} =
  \begin{cases}
    \mathbf{x}_i, & a_i=1, \\
    \Pi_{\mathrm{valid}}(\hat{\mathbf{z}}_{s,i}), & a_i=0.
  \end{cases}
  \label{eq:reverse_projection}
\end{equation}
Here $\hat{\mathbf{z}}_s$ is the model proposal before projection, and $\Pi_{\mathrm{valid}}$ enforces the validity rule in \cref{eq:valid_support} while preserving unresolved masks. This projection removes only invalid semantic--attribute combinations; it does not merge instances or change placement.   We analyze the token validity before and after projection during denoising in \cref{app:token_validity}.

During denoising, within each reverse step, semantic labels are updated first, then instance-color and facing predictions are recomputed from the updated semantic state. Labels that are already revealed are retained.
If the semantic label is masked, both instance-color and facing attributes remain masked.
When architecture/background is revealed,  the attributes are set to null.
To summarize, a mask value denotes an unknown value, whereas null denotes an inapplicable attribute.
This semantic-priority sampling occurs within one reverse process.  See \cref{app:inference} for details.

\subsection{Multichannel Denoising Network}
\label{sec:denoiser}

We train a multichannel denoising network $g_\theta$ to predict label probabilities for masked out tokens.
The model $g_\theta$ consists of a shared UNet $U_\theta$ which processes the summed field embeddings across height slices, and three prediction heads ($D_y,D_c,D_f$) for the semantic, instance-color, and facing
fields.
At each cell $i$, the three output heads produce the logits $\boldsymbol{\ell}_i^y,\boldsymbol{\ell}_i^c,\boldsymbol{\ell}_i^f$ that define the label probabilities used in \cref{sec:diffusion}.

\mypara{Joint RoomToken denoising.}
The main denoising step predicts the RoomToken values from a single shared hidden representation $\mathbf{h}_\theta$,
conditioned on the room type $r$ and arch mask $\mathbf{a}$.
For channel embeddings $E_y,E_c,E_f$, shared UNet $U_\theta$, and output heads $D_y,D_c,D_f$, the embedding token $\mathbf{e}_{t,i}$ at position $i$ for the denoised map at time $t$ is given by:
\begingroup
\setlength{\abovedisplayshortskip}{3pt}
\setlength{\belowdisplayskip}{5pt}
\begin{equation}
  \begin{gathered}
  \mathbf{e}_{t,i}=E_y(z^y_{t,i})+E_c(z^c_{t,i})+E_f(z^f_{t,i}), \quad \text{for each grid location } i, \\
  \mathbf{h}_{\theta}=U_{\theta}(\mathbf{e}_t;\,t,\kappa), \quad
  \boldsymbol{\ell}^{\rho}_i=D_\rho(\mathbf{h}_{\theta,i}), \quad \rho\in\{y,c,f\}.
  \end{gathered}
  \label{eq:multichannel_denoising}
\end{equation}
\endgroup
Here  $\mathbf{h}_{\theta,i}$ is the hidden feature, $\boldsymbol{\ell}^{\rho}_i$ the logits for channel $\rho$, and $\kappa=(r,\mathbf{a})$ the conditioning input. The clean $\mathbf{x}_i$ and noisy $\mathbf{z}_{t,i}$ variables are discrete categorical states; $\mathbf{e}_{t,i}$ and $\mathbf{h}_{\theta,i}$ are continuous vectors, and the three decoders map the shared vector to categorical channel logits.
Summed embeddings and a shared hidden state parameterize one RoomToken distribution rather than a semantic-then-attribute pipeline.

\mypara{Cross-slice consistency.}
Cross-slice attention provides information exchange across height slices at the same floor location.
The UNet treats the $H\times W$ grid as the spatial domain after concatenating $S$ slice embeddings per location. At attention resolutions, a cross-slice block reshapes each $(u,v)$ feature into $S$ tokens, applies attention along the slice dimension, and writes the result back through a residual projection. Thus, predictions for $\{(u,v,s):s=1,\ldots,S\}$ exchange evidence about object extent and body/front alignment.

\mypara{Cross-channel consistency.}
Cross-channel consistency makes instance-color and facing predictions depend on the semantic prediction. Instance-color and facing logits share the semantic backbone features, and the attribute branch is modulated by a semantic feature branch through learned gates and biases. This makes attributes semantic-aware within the same denoising step, while Sec.~\ref{sec:diffusion} enforces  channel combinations are valid during sampling.

\mypara{Training objective.}
The training objective supervises both token denoising and the geometric quantities needed for scene assembly. The main terms denoise the masked RoomToken channels, while auxiliary terms calibrate occupancy and the geometric heads used downstream. We optimize
\begin{equation}
  \mathcal{L} =
  \lambda_y\mathcal{L}_y + \lambda_{\mathrm{cls}}\mathcal{L}_{\mathrm{cls}}
  + \lambda_c\mathcal{L}_c + \lambda_f\mathcal{L}_f
  + \lambda_{\mathrm{occ}}\mathcal{L}_{\mathrm{occ}}
  + \lambda_{\mathrm{attr}}\mathcal{L}_{\mathrm{attr}} .
  \label{eq:total_loss}
\end{equation}
Here, each $\lambda$ is a scalar weight.
$\mathcal{L}_y$ is masked cross-entropy over semantic classes, while $\mathcal{L}_{\mathrm{cls}}$ is an object-count-balanced semantic class loss for the long-tailed furniture distribution. It applies fixed inverse-square-root weights computed from training object counts, keeps architecture/background weights at one, averages semantic cross-entropy over the cells of each ground-truth object, and then averages over objects. This gives rare categories and small instances comparable influence to common or large furniture. $\mathcal{L}_c$ and $\mathcal{L}_f$ are masked cross-entropy losses for instance-color and facing labels on visible furniture cells. $\mathcal{L}_{\mathrm{occ}}$ is binary cross-entropy with logits on foreground occupancy derived from the semantic logits. $\mathcal{L}_{\mathrm{attr}}$ groups instance-level mean-squared regression losses for yaw $(\cos\theta,\sin\theta)$, height, and vertical offset from instance-pooled features. See \cref{app:losses} for details.
\subsection{Inference and Scene Assembly}
\label{sec:inference}

At inference, architectural cells from the input arch mask are clamped throughout the reverse process, and \methodname{} samples semantic, instance-color, and facing slices for the furnished interior.
Because these architectural tokens are never overwritten,
the masked discrete diffusion design preserves the architectural condition throughout generation, enabling the model to learn furniture layouts within the valid room interior.  In our experiments (\cref{sec:exp_results}), we show that there are zero out-of-bounds errors after final asset placement.

We group connected RoomTokens with the same semantic and instance-color labels into object instances. Their tokens determine category, horizontal center, and horizontal dimensions; body/front labels indicate orientation. MLP heads within \methodname{} use denoising features averaged over each instance mask to predict continuous height and vertical offset, with a yaw head providing orientation when body/front labels do not determine it.
We instantiate the final scene by retrieving appropriate 3D assets from 3D-FUTURE~\citep{fu20213dfuture}.
For each object, we query the 3D database with the object category and select the closest asset based on the 3D extents.
The retrieved asset is scaled along each axis to match the estimated extents, rotated to the predicted yaw, and translated to the recovered object position.
We use the same category- and size-based asset retrieval and placement process as SemLayoutDiff. After asset placement, there is no clipping, extra optimization, or any hand-designed scene correction.
\Cref{app:inference} details assembly and the grouping/orientation procedures for the field-removal ablations.

%% file: figs/fig-representation.tex
\begin{figure}[t]
  \centering
  \includegraphics[trim={0 0 0 0},clip,width=\textwidth]{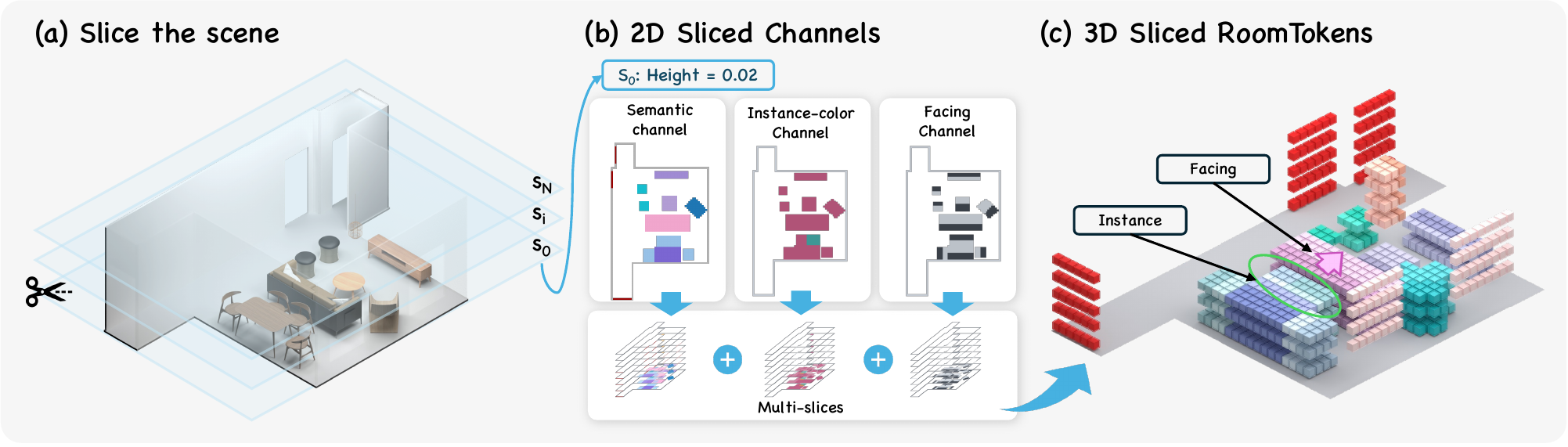}
  \caption{RoomToken data representation. (a) A 3D scene is represented as horizontal slices taken at selected vertical heights. (b) Each slice encodes three aligned 2D channels: semantic class, instance-color identity, and the facing field. Stacking these channels over all heights forms a sliced token volume. (c) The resulting 3D representation of stacked slices (RoomTokens) stores semantic, instance, and orientation information, with height derived from the slice index.
  \vspace{-0.25em}
  }
  \label{fig:representation}
\end{figure}

%% file: figs/fig-method.tex
\begin{figure}[t]
  \centering
  \includegraphics[trim={0 10pt 0 0pt},clip,width=\textwidth]{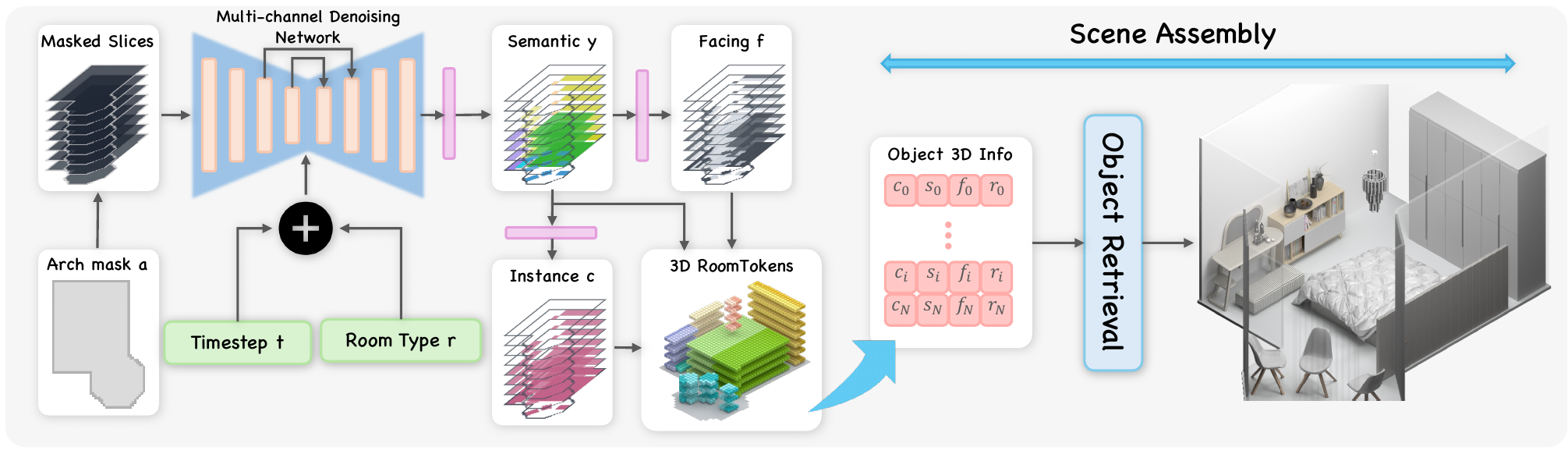}  \caption{Overview of \methodname. Given selected height slices and an architectural mask, we clamp the fixed room structure and denoise masked interior RoomTokens with a shared UNet with cross-slice attention. The denoiser predicts semantic, instance-color, and facing slices under semantic-gated validity constraints: architecture/background tokens carry no instance-color or facing cues, while furniture tokens do. The valid RoomToken grid is then grouped into object records, whose category, size, and pose are used to retrieve and place matched 3D assets.
  }
  \vspace{-1em}
  \label{fig:method}
\end{figure}

%% file: sections/04-experiments.tex
\section{Experiments}
\label{sec:exp}

We evaluate architecture-conditioned generation against recent diffusion methods.  Ablations and a user study assess design choices and perceived quality.

\subsection{Experimental Setup}
\label{sec:exp_setup}

\mypara{Dataset.}
We train and evaluate on 3D-FRONT~\citep{fu20213dfront} following splits established in prior work~\citep{paschalidou2021atiss} with 4{,}496 train / 274 val / 390 test rooms over three room types (bedroom, living room, and dining room).

\mypara{Model details.} Following SemLayoutDiff~\citep{sun2026semlayoutdiff}, we train all methods as a single unified model across room types rather than one model per room type. Each scene is rasterized on a $128{\times}128$ floor-plane grid, where one pixel/token has side length $0.10$ m in the horizontal plane. The grid is repeated over $S{=}8$ fixed height slices at $(0.02,0.45,0.80,1.20,1.60,2.00,2.40,2.70)$ m, which are selected once by greedy coverage analysis with a $0.10$ m vertical tolerance. For \methodname{}, RoomTokens use $K_{\mathrm{sem}}{=}39$ semantic classes, $C{=}3$ instance colors plus $\varnothing_c$, and body/front/$\varnothing_f$ facing labels; valid semantic--instance-color--facing triples are flattened into a 209-id palette for visualizations and the single-channel ablation. See \cref{app:model_details} for model details and \cref{app:representation_construction} for data processing details.

\mypara{Baselines.}
We compare \diffuscene~\citep{tang2024diffuscene} (continuous diffusion), \midiff~\citep{hu2026mixed} (mixed discrete-continuous diffusion), and \semdiff~\citep{sun2026semlayoutdiff} (semantic-map diffusion), all trained as unified models on the same 3D-FRONT split. We also retrain a \semdiff{} variant that retains multinomial diffusion while fixing exterior, door, and window pixels during training and sampling, restricting furniture to valid floor regions. We follow SemLayoutDiff's generation and evaluation procedures and reproduce their reported results for baselines.

\mypara{Metrics.}
Following prior work~\citep{sun2026semlayoutdiff}, we use FID and KID to assess rendered layout fidelity and CKL to compare object-category distributions against the 3D-FRONT test set. We introduce Pair-Rel to compare object-pair categories and spatial relations. For physical plausibility, we report out-of-boundary rates for scenes (OOB$_S$) and objects (OOB$_O$), object collisions (COL), navigability (NAV), and our door/window opening-clearance metric (OC). We also report the mean number of objects per scene (\#Obj.). Metric definitions and computation details are in \cref{app:metrics}. All scene-quality metrics are computed after asset retrieval, scaling, rotation, and placement, using the same 3D-FUTURE assets, SceneToolkit rendering, and evaluation scripts as SemLayoutDiff. No subsequent clipping, optimization, or hand-designed scene correction is applied.

\subsection{Results and Analysis}
\label{sec:exp_results}

\paragraph{Physical validity with competitive layout quality.}
\label{sec:exp_quantitative}

\input{tables/table-comparison}
We compare output from \methodname{} against baselines with scenes generated from five different randomly sampled seeds.
Table~\ref{tab:comparison} shows that \methodname{} achieves a better FID than SemLayoutDiff (50.14 vs.~71.18) and comparable KID/CKL (10.35/6.81 vs.~8.52/4.85), demonstrating competitive distribution learning. It also performs best across all spatial plausibility metrics: lower Pair-Rel (12.09) indicates more plausible object relations, zero OOB after asset placement confirms boundary compliance, lower COL (12.64) reflects fewer intersections, higher NAV (99.25) indicates better free-space connectivity, and higher OC (82.31) indicates better clearance around openings. The average  of 8.25 objects per scene for \methodname{},  compared with 7.96 for SemLayoutDiff and 9.39 in the ground truth, confirms that the gains  are not due to sparse layouts.

FID and KID compare rendered feature distributions using different statistics; CKL compares category frequencies. None directly measures 3D collisions, opening clearance, or quality of the scene. We therefore also assess object relations, physical plausibility, and human preference.

SemLayoutDiff with architecture clamping reduces OOB but raises COL from 16.24 to 40.89 because its reverse process denoises random semantic values rather than masked furniture under a fixed architecture. Clamping also cannot account for geometry added during grouping, retrieval, and rotation, so some objects remain partially outside the room. By jointly modeling arch-conditioned semantic, instance-color, and facing slices, \methodname{} achieves zero OOB while retaining low collision with small variation in COL and OC across runs.

\paragraph{\methodname{} reduces visible layout failures.}
\label{sec:exp_qualitative}
Figure~\ref{fig:qualitative} shows that the quantitative improvements are visible in the generated layouts: baseline methods can place objects across boundaries, collide nearby furniture, or block architectural openings such as doors and windows, while \methodname{} better preserves clear openings, room geometry, and object structure.  See \cref{app:additional_qualitative_results} for additional results.

\input{figs/fig-qualitative}

\input{tables/table-ablation}

\mypara{\textsc{Ablation and failure analysis.}}
\label{sec:exp_ablation}
We conduct ablation experiments on our model to determine the impact of different components (\cref{tab:ablation,tab:ablation_design}).
All ablation models are trained from scratch with the same data split and training settings, changing only the specified component. All use the architectural condition and the same scene assembly, rendering, and evaluation procedure.

\mypara{Class balancing and multichannel prediction.}
Without class balancing ($\mathcal{L}_\text{cls}$), object count falls from 8.36 to 6.31 and CKL rises from 6.43 to 21.31, yielding sparser scenes with less accurate category frequencies. Combining the three fields into one categorical token, with all eight slices retained, raises FID from 49.84 to 95.35 and Pair-Rel from 12.13 to 48.32, supporting multichannel prediction.

\mypara{Instance color and facing.}
Removing the instance-color field raises COL to 88.77, indicating less reliable separation of object instances. Removing the facing field raises COL to 89.07 and lowers NAV from 99.30 to 75.69, indicating more overlap and less navigable scenes. Without instance color, connected same-category tokens form one instance.
Without facing, an auxiliary head predicts the orientation from instance features (\cref{app:inference}).
These comparisons support retaining instance separation and orientation cues.

\mypara{Cross-slice attention and sampling.}
Removing cross-slice attention raises COL from 12.75 to 89.01, showing its importance for avoiding object intersections. Without semantic-priority sampling, COL rises to 58.81 and OC falls to 54.90, supporting the prediction of categories before instance-color and facing fields.

\input{tables/table-ablation-design}

\label{sec:exp_design_ablation}
\mypara{Multi-slice vs.~single-slice RoomTokens.}
The single-slice variant projects the scene into one top-down map. Multi-slice RoomTokens generate more objects (8.36 vs.~6.08) yet achieve higher OC (82.04 vs.~74.67) and lower COL (12.75 vs.~13.75). Thus, retaining vertical information improves physical plausibility without relying on sparser scenes.

\mypara{RoomTokens versus semantic maps.}
With masked diffusion and clamping fixed, multi-slice RoomTokens have lower COL (12.75 vs.~19.50) and higher OC (82.04 vs.~74.97) than top-down semantic maps using SemLayoutDiff's attribute prediction model (APM), despite generating more objects (8.36 vs.~7.03). This supports encoding instance color, facing, and vertical information for physical plausibility.

\mypara{Three versus four instance colors.}
Four instance colors expand valid semantic--instance-color--facing combinations from 209 to 277, adding embedding entries and output channels for combinations unused when three colors suffice. Four colors improve Pair-Rel but worsen COL (12.75 to 17.49) and OC (82.04 to 67.74), supporting the three-color choice. \Cref{app:data_analysis} provides instance-separation diagnostics.

\mypara{Diffusion and architectural clamping.}
With clamping, masked diffusion has lower COL (12.75 vs.~32.16) and higher OC (82.04 vs.~62.53) than multinomial diffusion. Removing clamping raises scene OOB from 0.00\% to 1.17\% for masked diffusion and from 0.03\% to 34.60\% for multinomial diffusion. Clamping reduces OOB in both models but raises collisions for multinomial diffusion; masked diffusion with clamping achieves zero OOB and the lowest COL.

\mypara{Failure cases.} Figure~\ref{fig:failure_cases} shows representative failures: implausible furniture arrangements, object--object intersections, and blocked doorways. Valid RoomTokens and in-boundary placement do not guarantee functional arrangements or sufficient clearance in the generated scene.

\noindent
\begin{minipage}[t]{0.55\textwidth}
\textbf{User study.}
We ask participants to rank scenes based on architectural fidelity and feasibility of object arrangements. Each participant is shown 20 sets of four generated scenes from different methods.
The scenes are randomized, and users are asked to rank the scenes from best to  worst.
We had 12 participants in the user study.
Overall, participants ranked \methodname{} best on average (Table~\ref{tab:user_study}).
This preference is significant against every baseline with paired Wilcoxon tests after Holm correction ($p{=}0.0015$),
and indicates that participants find \methodname{} to have stronger perceived architectural fidelity and object arrangement than baselines.
 The paired Wilcoxon tests account for within subject repeated ratings. See \cref{app:additional_experiments} for user study details.

\end{minipage}\hfill
\input{tables/table-user-study}
\par

\input{figs/fig-failure-cases}

%% file: tables/table-comparison.tex
\begin{table}
  \centering
  \caption{Quantitative comparison on 3D-FRONT. Results are mean $\pm$ standard deviation across 5 randomly sampled seeds during inference, with each trained model fixed. KID is reported as KID$\times10^3$ and CKL as CKL$\times10^2$. Best mean values are bold and second-best values are italic.}
  \label{tab:comparison}
  \setlength{\tabcolsep}{3pt}
  \resizebox{\textwidth}{!}{
  \begin{tabular}{l cccc ccccc c}
    \toprule
    & \multicolumn{5}{c}{Distribution / Composition} & \multicolumn{5}{c}{Spatial / Physical Plausibility} \\
    \cmidrule(lr){2-6} \cmidrule(lr){7-11}
    Method & \fid & \kid & \ckl & Pair-Rel$\downarrow$ & \#Obj. & \oobscene & \oobobj & \col & \nav & OC$\uparrow$ \\
    \midrule
    \diffuscene~\cite{tang2024diffuscene} & 88.52\,{\scriptsize$\pm$\,0.35} & 30.15\,{\scriptsize$\pm$\,0.42} & 20.49\,{\scriptsize$\pm$\,0.28} & 32.38\,{\scriptsize$\pm$\,0.95} & 8.71\,{\scriptsize$\pm$\,0.15} & 66.35\,{\scriptsize$\pm$\,0.48} & 20.15\,{\scriptsize$\pm$\,0.31} & 41.06\,{\scriptsize$\pm$\,0.44} & 94.22\,{\scriptsize$\pm$\,0.25} & 59.58\,{\scriptsize$\pm$\,1.05} \\
    \midiff~\citep{hu2026mixed} & 93.63\,{\scriptsize$\pm$\,0.41} & 30.95\,{\scriptsize$\pm$\,0.37} & 37.05\,{\scriptsize$\pm$\,0.22} & 40.52\,{\scriptsize$\pm$\,0.46} & 7.42\,{\scriptsize$\pm$\,0.18} & 60.55\,{\scriptsize$\pm$\,0.39} & 34.78\,{\scriptsize$\pm$\,0.98} & 51.51\,{\scriptsize$\pm$\,0.29} & 96.28\,{\scriptsize$\pm$\,0.33} & \textit{75.82\,{\scriptsize$\pm$\,0.47}} \\
    \semdiff~\citep{sun2026semlayoutdiff} & \textit{71.18\,{\scriptsize$\pm$\,0.34}} & \textbf{8.52\,{\scriptsize$\pm$\,0.26}} & \textit{4.85\,{\scriptsize$\pm$\,0.19}} & \textit{20.65\,{\scriptsize$\pm$\,0.43}} & 7.96\,{\scriptsize$\pm$\,0.12} & 16.71\,{\scriptsize$\pm$\,0.38} & 6.55\,{\scriptsize$\pm$\,0.21} & \textit{16.24\,{\scriptsize$\pm$\,0.92}} & \textit{96.45\,{\scriptsize$\pm$\,0.27}} & 56.02\,{\scriptsize$\pm$\,0.45} \\
    \semdiff{} (arch. clamped) & 73.45\,{\scriptsize$\pm$\,0.36} & 10.63\,{\scriptsize$\pm$\,0.24} & \textbf{4.51\,{\scriptsize$\pm$\,0.17}} & 22.15\,{\scriptsize$\pm$\,0.41} & 7.32\,{\scriptsize$\pm$\,0.14} & \textit{0.41\,{\scriptsize$\pm$\,0.08}} & \textit{0.35\,{\scriptsize$\pm$\,0.05}} & 40.89\,{\scriptsize$\pm$\,0.49} & 95.25\,{\scriptsize$\pm$\,0.31} & 65.48\,{\scriptsize$\pm$\,1.02} \\
    \midrule
    \textbf{Ours} & \textbf{50.14\,{\scriptsize$\pm$\,0.27}} & \textit{10.35\,{\scriptsize$\pm$\,0.21}} & 6.81\,{\scriptsize$\pm$\,0.66} & \textbf{12.09\,{\scriptsize$\pm$\,0.07}} & 8.25\,{\scriptsize$\pm$\,0.18} & \textbf{0.00\,{\scriptsize$\pm$\,0.00}} & \textbf{0.00\,{\scriptsize$\pm$\,0.00}} & \textbf{12.64\,{\scriptsize$\pm$\,0.34}} & \textbf{99.25\,{\scriptsize$\pm$\,0.24}} & \textbf{82.31\,{\scriptsize$\pm$\,0.34}} \\
    \bottomrule
  \end{tabular}
  }
\end{table}

%% file: figs/fig-qualitative.tex
\begin{figure}[t]
  \centering  \includegraphics[trim={0 5pt 0 5pt},clip,width=\columnwidth]{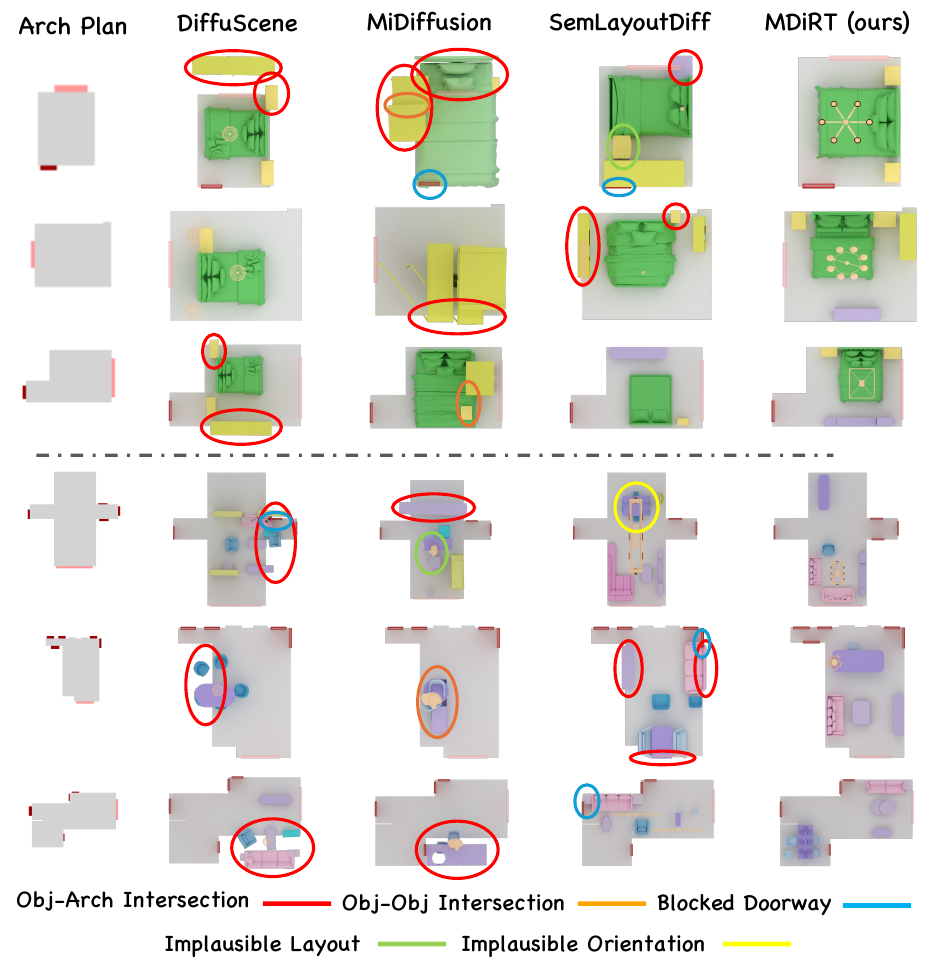}
  \vspace{-1.5em}  \caption{Arch-mask-conditioned qualitative comparison on 3D-FRONT. We use semantic orthographic views to focus the comparison on layout evaluation. Baselines show boundary violations, collisions, or blocked door/window openings, while \methodname{} preserves clear openings, room geometry, and structured object cues.
  }
  \vspace{-1em}
  \label{fig:qualitative}
\end{figure}

%% file: tables/table-ablation.tex
\begin{table*}[t]
  \centering
  \caption{
  Ablations for \methodname{}.
  Best values are in bold. %
  }
  \label{tab:ablation}
  \resizebox{\textwidth}{!}{
  \begin{tabular}{l cccc ccccc c}
    \toprule
    & \multicolumn{5}{c}{Distribution / Composition} & \multicolumn{5}{c}{Spatial / Physical Plausibility} \\
    \cmidrule(lr){2-6} \cmidrule(lr){7-11}
    Configuration & \fid & \kid & \ckl & Pair-Rel$\downarrow$ & \#Obj. & \oobscene & \oobobj & \col & \nav & OC$\uparrow$ \\
    \midrule
    Multichannel training (\methodname{})                      & \textbf{49.84} & \textbf{10.41} & \textbf{6.43} & \textbf{12.13} & 8.36 & \textbf{0.00} & \textbf{0.00} & \textbf{12.75} & \textbf{99.30} & \textbf{82.04} \\
    \ w/o class balancing                               & 74.77 & 34.18 & 21.31 & 20.03 & 6.31 & \textbf{0.00} & \textbf{0.00} & 37.73 & 96.58 & 76.20 \\
    \ w/o multichannel (single-channel)                  & 95.35 & 34.91 & 30.05 & 48.32 & 7.12 & \textbf{0.00} & \textbf{0.00} & 49.30 & 95.74 & 66.40 \\
    \ w/o semantic-priority sampling           & 88.21 & 28.74 & 32.40 & 34.54 & 7.84 & \textbf{0.00} & \textbf{0.00} & 58.81 & 73.07 & 54.90 \\
    \ w/o instance-color field                                   & 98.73 & 36.42 & 51.96 & 37.32 & 9.41 & \textbf{0.00} & \textbf{0.00} & 88.77 & 83.52 & 48.30 \\
    \ w/o facing field                                            & 84.92 & 26.18 & 28.62 & 36.63 & 8.96 & \textbf{0.00} & \textbf{0.00} & 89.07 & 75.69 & 50.70 \\
    \ w/o cross-slice attention                                  & 91.05 & 31.27 & 37.82 & 34.78 & 9.18 & \textbf{0.00} & \textbf{0.00} & 89.01 & 87.16 & 55.60 \\
    \bottomrule
    \vspace{-1em}
  \end{tabular}
  }
\end{table*}

%% file: tables/table-ablation-design.tex
\begin{table}[t]
  \centering
  \caption{\methodname{} ablations of representation, instance colors, diffusion, and architectural clamping. KID and CKL are scaled by $10^3$ and $10^2$. Best values are bold.}
  \vspace{-0.5em}
  \label{tab:ablation_design}
  \setlength{\tabcolsep}{3pt}
  \resizebox{\textwidth}{!}{
  \begin{tabular}{lclcrrrrrrrrrr}
    \toprule
    \multicolumn{4}{c}{Variant settings} & \multicolumn{5}{c}{Distribution / Composition} & \multicolumn{5}{c}{Spatial / Physical Plausibility} \\
    \cmidrule(lr){1-4} \cmidrule(lr){5-9} \cmidrule(lr){10-14}
    Representation & \makecell{Instance\\colors} & Diffusion & Clamping & \fid & \kid & \ckl & Pair-Rel$\downarrow$ & \#Obj. & \oobscene & \oobobj & \col & \nav & OC$\uparrow$ \\
    \midrule
    Multi-slice (full) & 3 & Masked & On & \textbf{49.84} & 10.41 & 6.43 & 12.13 & 8.36 & \textbf{0.00} & \textbf{0.00} & \textbf{12.75} & \textbf{99.30} & \textbf{82.04} \\
    Single-slice (top-down) & 3 & Masked & On & 62.57 & 18.64 & 11.11 & 9.02 & 6.08 & \textbf{0.00} & \textbf{0.00} & 13.75 & 97.35 & 74.67 \\
    Top-down semantic map & -- & Masked & On & 54.69 & \textbf{8.89} & 10.15 & 13.97 & 7.03 & \textbf{0.00} & \textbf{0.00} & 19.50 & 97.59 & 74.97 \\
    Multi-slice & 4 & Masked & On & 58.03 & 14.06 & 9.74 & \textbf{8.40} & 6.67 & \textbf{0.00} & \textbf{0.00} & 17.49 & 97.26 & 67.74 \\
    Multi-slice & 3 & Masked & Off & 57.61 & 14.02 & 8.71 & 8.98 & 6.59 & 1.17 & 0.18 & 17.64 & 97.44 & 70.94 \\
    Multi-slice & 3 & Multinomial & On & 70.57 & 16.14 & \textbf{5.32} & 11.02 & 8.00 & 0.03 & 0.0042 & 32.16 & 93.10 & 62.53 \\
    Multi-slice & 3 & Multinomial & Off & 70.14 & 18.59 & 12.92 & 16.78 & 8.67 & 34.60 & 15.88 & 17.74 & 96.09 & 69.65 \\
    \bottomrule
  \end{tabular}
  }
\end{table}

%% file: tables/table-user-study.tex
\begin{minipage}[t]{0.40\textwidth}
 \begingroup
 \centering
  \captionof{table}{User study with 12 participants ranking generated scenes.  \methodname{} was ranked 1st almost $80\%$ of the time.
  }
  \label{tab:user_study}
  \scriptsize
  \begin{tabular*}{\linewidth}{@{\extracolsep{\fill}}lccc@{}}
    \toprule
    Method & Mean rank$\downarrow$ & First$\uparrow$ & Top-2$\uparrow$ \\
    \midrule
    \textbf{\methodname{} (Ours)} & \textbf{1.25} & \textbf{79.6\%} & \textbf{96.3\%} \\
    \semdiff{} & 2.15 & 18.3\% & 73.3\% \\
    \diffuscene{} & 3.15 & 2.1\% & 17.9\% \\
    \midiff{} & 3.45 & 0.0\% & 12.5\% \\
    \bottomrule
  \end{tabular*}
 \endgroup
\end{minipage}

%% file: figs/fig-failure-cases.tex
\begin{figure}[tbh]
  \centering
  \includegraphics[trim={0 10pt 0 5pt},clip,width=\columnwidth]{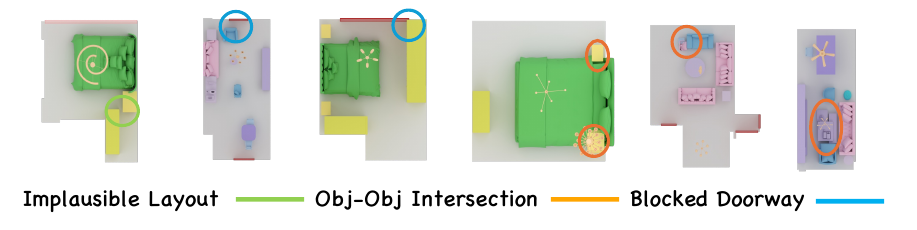}
  \captionof{figure}{Failure cases: implausible layouts (green), object intersections (orange), and blocked doorways (blue).}
  \label{fig:failure_cases}
  \vspace{-1em}
\end{figure}

%% file: sections/05-conclusion.tex
\section{Conclusion}
\label{sec:conclusion}

We presented \methodname{}, a masked discrete diffusion framework over sliced RoomTokens for indoor scene layouts. RoomTokens add height slices, instance-color cues, and facing cues to semantic occupancy maps. Clamping architectural tokens during denoising preserves the supplied floor plan, keeping room geometry explicit while supporting object grouping, orientation estimation, and scene assembly.

On 3D-FRONT, \methodname{} achieves competitive distribution quality and the strongest overall spatial plausibility among evaluated methods. Controlled comparisons show gains beyond architecture clamping and sparse generation; a blinded human ranking study favors our scenes. Representation checks show little information loss; ablations support factorized RoomToken fields, semantic-priority sampling, and cross-slice attention. Structured occupancy tokens could also support outdoor scenes, objects with parts, and other semantic occupancy tasks.

\mypara{Limitations.}
The fixed grid and height slices can miss fine details and narrow gaps. The small 3D-FRONT training set is uneven across object categories and room types. RoomTokens and validity projection cannot prevent mesh collisions from grouping, continuous pose estimation, scaling, or asset retrieval. Larger, more diverse data and stronger scene assembly could improve distribution matching and retrieval; adaptive-resolution slices and geometry-aware refinement could capture details and reduce collisions.

%% file: sections/acknowledgments.tex
\section*{Acknowledgments}
\label{sec:ack}

This work was funded in part by a CIFAR AI Chair and NSERC Discovery Grants, and enabled by support from the \href{https://alliancecan.ca/}{Digital Research Alliance of Canada} and a CFI/BCKDF JELF. We thank Jiayi Liu and Yiming Zhang for constructive feedback and discussion.

%% file: sections/06-appendix.tex
\section*{Appendix}
\appendix
\noindent In this appendix, we provide the evaluation metric definitions (\cref{app:metrics}),
additional experiments (\cref{app:additional_experiments}),
diagnostics of the dataset (\cref{app:data_analysis}), implementation details for our model (\cref{app:implementation_details}), details of RoomTokens construction (\cref{app:representation_construction}), final scene assembly details (\cref{app:inference}), and additional qualitative results (\cref{app:additional_qualitative_results}).

\section{Evaluation Metrics}
\label{app:metrics}

We follow prior work~\citep{sun2026semlayoutdiff} and evaluate the distribution similarity of generated scenes to reference scenes, as well as physical layout plausibility with metrics introduced in SceneEval~\citep{tam2026sceneeval}.

\mypara{Distribution and semantic quality.}
For FID and KID, we render each generated scene and each 3D-FRONT test scene as a $256{\times}256$ orthographic top-down image. Objects are colored by semantic category, using the same fixed room-type palette for generated and test scenes. We convert the renders to RGB and white-pad them to $256{\times}256$. We extract clean-fid ``clean'' Inception features and compute FID and KID from these features. For each comparison, we use equal-sized image sets, sample without replacement from the larger set, and average scores over ten draws. The aggregate \methodname{} score compares all 390 test renders with 390 renders drawn from 3,000 generated scenes on each draw; per-room scores are computed separately. KID is reported multiplied by $10^3$.

CKL is $D_{\mathrm{KL}}(p_{\mathrm{gt}}\Vert p_{\mathrm{gen}})$ between normalized furniture-category counts pooled across test and generated scenes, excluding architectural elements. CKL is reported multiplied by $10^2$.

\mypara{Pairwise layout relations.}
Our Pair-Rel metric compares the joint distributions of object-pair categories and relative geometry. For every ordered pair $(i,j)$ of distinct furniture objects with valid bounding boxes, we form a relation tuple
$r_{ij}=(k_i,k_j,d_{ij},b_{ij})$, where $k_i$ and $k_j$ are the object categories, $d_{ij}$ is the direction of object $j$ relative to object $i$, and $b_{ij}$ is a distance bin. Both $(i,j)$ and $(j,i)$ contribute.

To compute $d_{ij}$, we express the displacement between object centers in object $i$'s orientation frame. The axis with the larger displacement magnitude determines left, right, front, or back; coincident centers receive the label same. We measure distance as the minimum horizontal separation between the objects' oriented bounding boxes, assigning zero to intersecting boxes. The distance bins are contact ($\leq 0.05$\,m), near ($(0.05,0.5]$\,m), mid ($(0.5,1.5]$\,m), and far ($>1.5$\,m).

We pool relation tuples across all scenes in each set and normalize their counts to obtain generated and ground-truth histograms $P_{\mathrm{gen}}$ and $P_{\mathrm{gt}}$.
Let $M=\tfrac12(P_{\mathrm{gen}}+P_{\mathrm{gt}})$. Pair-Rel is
\begin{equation}
\begin{gathered}
\operatorname{PairRel}
=100\,\operatorname{JSD}(P_{\mathrm{gen}},P_{\mathrm{gt}})\\
=50\left[
D_{\mathrm{KL}}(P_{\mathrm{gen}}\Vert M)
+D_{\mathrm{KL}}(P_{\mathrm{gt}}\Vert M)
\right].
\end{gathered}
\end{equation}
Lower values indicate closer agreement. Because object categories are included in each tuple, Pair-Rel reflects both pair composition and relative geometry.

\mypara{Physical plausibility.}
OOB$_S$ is the fraction of generated scenes containing at least one object outside the room boundary. OOB$_O$ is the fraction of generated objects outside the room boundary. COL measures collisions among generated objects after connected-component grouping, orientation estimation, and object placement. NAV evaluates free-space navigability in the final scene. Opening clearance (OC) is the percentage of evaluated room-facing door and window clearance regions that do not intersect a furniture mesh. Mean object count is reported alongside physical metrics to expose empty or unusually sparse generation.  All metrics use the final scenes and evaluation protocol described in \cref{sec:exp_setup}.

\section{Additional Experiments and Details}
\label{app:additional_experiments}

We provide unconditional and per-room results, denoising diagnostics, and user-study analysis.

\input{tables/table-unconditional}

\mypara{Unconditional generation.}
For unconditional generation, we use a separate room-type-conditioned multinomial diffusion model to first generate a $128{\times}128$ architecture map.
The architectural map is provided as input to \methodname{}, which lifts it to the eight RoomToken slices and passes it to the frozen layout generator. Table~\ref{tab:unconditional} shows that this two-stage pipeline (\textbf{ArchGen $\rightarrow$ MDiRT}) improves FID, COL, and NAV over unconditional SemLayoutDiff while retaining zero OOB.

\mypara{Per-room-type results.}
\input{tables/table-per-room}
We report per room-type evaluation results in \cref{tab:per_room}. \methodname{} retains zero OOB for every room type and has the lowest collision rate for bedrooms and living rooms, while SemLayoutDiff has lower collisions for dining rooms. \methodname{} has the best clearance around openings (highest OC).
It is able to achieve reasonable plausibility metrics at a comparable object rate per room-type as other methods.
Distributional performance is more mixed: \methodname{} has the lowest FID and Pair-Rel, whereas SemLayoutDiff remains slightly stronger on most per-room KID/CKL results.

 \mypara{Token validity before projection.}
\label{app:token_validity}
We measure token validity before and after $\Pi_\text{valid}$ during denoising to determine how often the model predicts valid field combinations without correction.
We record all 256 reverse steps for the full 390-condition test set with one seed per condition. Before validity projection, 99.49\% of all evaluated tokens are valid at the final step and furniture-only tokens are 100\% valid. After projection, validity is 100\%. The central semantic reveal interval spans approximately steps 74--240, 95\% of final occupancy is present by step 248, and 99\% of complete semantic--instance-color--facing triples are present by step 256.  Thus, the denoiser already produces compatible field combinations for almost all tokens, including all furniture tokens, before the final  projection.

\mypara{Is the model memorizing?}
\input{tables/table-novelty}
We also check whether our model is capable of generating new layout by retrieving the closest sample from the training set.
The experiment uses 100 conditions
from each of three room types and 10 seeds per condition.
The results in \cref{tab:novelty} provide evidence against direct memorization but do not prove novelty: 12 of the 300 conditions trigger a strict exact-duplicate heuristic for at least one sample.

\mypara{User-study details.}
The user study consisted of 12 participants who provided 240 total rankings. Each participant evaluates 20  scene sets: 10 bedrooms and 10 living/dining rooms.  Participants view different randomly selected sets, covering 50 scene sets across three room types. For each set, we show four anonymized scenes in random order, with each method appearing equally often in each display position. Participants rank the scenes from best to worst based on architectural fidelity and object arrangement, with no ties.

To account for repeated ratings, we average each participant's rankings for each method and apply statistical tests to these participant averages. The Friedman test gives $\chi^2(3)=34.0$ ($p<2\times10^{-7}$). Paired Wilcoxon tests on these participant means remain significant against every baseline after Holm correction ($p=0.0015$). All 12 participants prefer \methodname{} on average. \Cref{tab:user_study_per_room} reports descriptive results by room type: \methodname{} has the best mean rank in each type.
\input{tables/table-user-study-per-room}

\section{Dataset and Representation Diagnostics}
\label{app:data_analysis}

\mypara{Representation sanity checks.}
\label{sec:exp_representation_stats}
We conduct a set of representation checks to verify that RoomTokens preserve object visibility, instance separation, and orientation before model evaluation. The processed data keeps 99.47\% of furniture boxes visible, has 0.00\% invalid tokens, only 0.92\% same-class color conflicts, and 1.13$^{\circ}$ direct orientation MAE, indicating little representation loss.

\mypara{RoomToken diagnostics.}
These metrics evaluate the RoomTokens representation and are not used to compare methods.

Table~\ref{tab:representation_stats_app} reports dataset-level statistics of the processed 3D-FRONT RoomToken representation before model training. Object coverage measures the fraction of furniture boxes that appear in at least one rasterized height slice. Invalid token counts semantic--instance-color--facing triples that violate the representation rules. Color conflict counts adjacent same-class, different-instance token pairs assigned the same graph color. Instance count MAE compares connected components recovered from $(y_i,c_i)$ with ground-truth instance counts per scene. B/F paired is the fraction of recovered components with both body and front cells, and orientation MAE measures the yaw error obtained by decoding body/front geometry directly.

\input{tables/table-representation-stats}

\mypara{Instance-color diagnostics.}
\label{app:palette_comparison}
With three instance colors, 99.08\% of adjacent token pairs belonging to different instances of the same category have different colors. Grouping tokens by semantic and instance-color labels recovers the correct instance count in 91.53\% of scenes, with an instance-count mean absolute error (MAE) of 0.18. These data-analysis results show few color conflicts in the processed representation; they do not establish universal three-color sufficiency.
In \Cref{tab:ablation_design}, we compare generation with three and four instance colors and show experimentally that while using four colors gives a lower Pair-Rel score, using three colors gives a better collision rate and opening clearance.

\mypara{Collision sources in the training data.}
We audit furniture--furniture collisions in the raw 3D-FRONT train and test room records. For each furniture object, we load its 3D-FUTURE asset mesh and apply its recorded scene transform. We test each unordered pair of meshes for intersection using the Flexible Collision Library (FCL) through \texttt{trimesh} and \texttt{python-fcl}; architectural elements are excluded. When usable contact points are available, we move one mesh by $5$\,mm toward its centroid from the mean contact point and count the pair only if the collision persists. Otherwise, we retain the initial intersection result. A room record is colliding if it contains at least one counted pair, and an object is colliding if it belongs to at least one pair.

In the training records, 2{,}112 of 4{,}882 rooms (43.26\%) contain a collision, and 6{,}279 of 31{,}256 furniture objects (20.09\%) participate in one. The 4{,}096 collision pairs count distinct unordered pairs within room records; one object can belong to multiple pairs. In the test records, 177 of 390 rooms (45.38\%) and 840 of 4{,}965 objects (16.92\%) are colliding, across 592 pairs. These raw-record counts differ from the filtered split used to train and evaluate the model. They show that furniture intersections are present in the source data; further intersections can arise when generated objects are reconstructed and placed.

 \section{Implementation Details}
\label{app:implementation_details}

\subsection{Model Details}
\label{app:model_details}

Models are trained only on the training split; the validation split is used for hyperparameter and checkpoint selection, and the held-out test set is used only for final evaluation. \methodname{} learns a distribution over structured RoomTokens with a multichannel training parameterization. The denoiser uses the same UNet backbone as the flat-token model, with channel multipliers $(1,1,2,2)$ and self-attention plus cross-slice attention at $32{\times}32$ and $16{\times}16$. Unless otherwise stated, models train for 3000 epochs on 2 NVIDIA L40S GPUs with batch size 24 per GPU, AdamW, learning rate $3\!\times\!10^{-4}$, and weight decay $0.01$, and are sampled with 256 reverse-diffusion steps. The structured-token loss uses masked semantic loss with object-count-balanced class weighting, instance-color/facing losses, occupancy calibration, and yaw/height/offset heads.

\subsection{Training Losses}
\label{app:losses}

Let $\mathcal M_b^y$ contain the masked semantic cells in scene $b$.
The sets $\mathcal M_b^c$ and $\mathcal M_b^f$ contain masked instance-color and
facing cells whose furniture semantic label is visible. With
$\operatorname{CE}$ denoting categorical cross-entropy, the three token
losses are
\begin{equation}
\mathcal L_\rho =
\frac{1}{B}\sum_{b=1}^{B}
\sum_{i\in\mathcal M_b^\rho}
\operatorname{CE}(\boldsymbol\ell_{b,i}^\rho,x_{b,i}^\rho),
\qquad \rho\in\{y,c,f\}.
\end{equation}

For the object-count-balanced semantic term, let $\mathcal O_b$ be the
ground-truth furniture objects with masked semantic cells, and let
$\mathcal S_{b,o}$ be those cells belonging to object $o$. We use
\begin{equation}
\mathcal L_{\mathrm{cls}} =
\frac{1}{B}\sum_{b=1}^{B}
\frac{1}{|\mathcal O_b|}
\sum_{o\in\mathcal O_b}
\frac{w_{k(o)}}{|\mathcal S_{b,o}|}
\sum_{i\in\mathcal S_{b,o}}
\operatorname{CE}(\boldsymbol\ell^y_{b,i},y_{b,i}),
\end{equation}
where scenes with $\mathcal O_b=\varnothing$ contribute zero,
$k(o)$ is the class of object $o$, and the fixed furniture-class
weights satisfy $w_k\propto(n_k+\epsilon)^{-1/2}$ for training-set
object count $n_k$. Averaging within each object before averaging
across objects prevents large instances from dominating this term.

Let $Y_{\mathrm{furn}}$ and $Y_{\mathrm{arch}}$ be the furniture and
architecture/background classes. The foreground logit and target are
\begin{equation}
g_{b,i} =
\operatorname{LSE}_{k\in Y_{\mathrm{furn}}}\ell^y_{b,i,k}
-
\operatorname{LSE}_{k\in Y_{\mathrm{arch}}}\ell^y_{b,i,k},
\qquad
q_{b,i}=\mathbf{1}[y_{b,i}\in Y_{\mathrm{furn}}],
\end{equation}
where $\operatorname{LSE}$ denotes log-sum-exp. The occupancy loss is
\begin{equation}
\mathcal L_{\mathrm{occ}} =
\frac{1}{B}\sum_{b=1}^{B}
\sum_{i\in\mathcal M_b^y}
\operatorname{BCEWithLogits}(g_{b,i},q_{b,i}).
\end{equation}

Finally, instance-pooled features predict yaw, height, and vertical
offset. Writing $\mathbf v_o=(\cos\theta_o,\sin\theta_o)$, we group
their regression losses as
\begin{equation}
\mathcal L_{\mathrm{attr}} =
\beta_\theta\operatorname{MSE}(\hat{\mathbf v},\mathbf v)
+\beta_h\operatorname{MSE}(\hat h,h)
+\beta_\delta\operatorname{MSE}(\hat\delta,\delta),
\end{equation}
where each MSE is averaged over supervised furniture objects and the
$\beta$ values weight the three attribute heads.

\section{RoomTokens Construction Details}
\label{app:representation_construction}

\mypara{Greedy slice-height selection.}
The slice heights are chosen to cover as many furniture vertical intervals as possible with a fixed number of horizontal planes. For each object $n$, let $I_n=[z^-_n,z^+_n]$ be its vertical extent. A candidate height $h$ covers object $n$ if $[h-\tau,h+\tau]\cap I_n\neq\emptyset$, where $\tau{=}0.10$m is the rasterization tolerance used in our preprocessing. \Cref{alg:slice_heights} samples candidate heights at 0.01m spacing, starts from fixed floor and ceiling anchors, and repeatedly adds the candidate height that covers the largest number of currently uncovered objects while staying at least $\delta{=}0.10$m from already selected heights.

\begin{algorithm}[H]
\caption{Greedy slice-height selection}
\label{alg:slice_heights}
\KwIn{Object height intervals $\mathcal{I}=\{I_n=[z^-_n,z^+_n]\}_{n=1}^N$; candidate heights $\mathcal{H}$; number of slices $S$; tolerance $\tau$; minimum spacing $\delta$; fixed anchors $\mathcal{A}$}
\KwOut{Selected slice heights $\mathcal{Z}$}
$\mathcal{Z}\leftarrow\mathcal{A}$\;
$\mathcal{U}\leftarrow\{1,\ldots,N\}$\tcp*{objects not yet covered}
Remove from $\mathcal{U}$ any object covered by a height in $\mathcal{Z}$\;
\While{$|\mathcal{Z}|<S$}{
  $\mathcal{H}_{\mathrm{ok}}\leftarrow\{h\in\mathcal{H}: |h-z|\geq\delta\text{ for all }z\in\mathcal{Z}\}$\;
  $g(h)\leftarrow\left|\{n\in\mathcal{U}: [h-\tau,h+\tau]\cap I_n\neq\emptyset\}\right|$ for each $h\in\mathcal{H}_{\mathrm{ok}}$\;
  $h^\star\leftarrow\arg\max_{h\in\mathcal{H}_{\mathrm{ok}}} g(h)$\;
  $\mathcal{Z}\leftarrow\mathcal{Z}\cup\{h^\star\}$\;
  Remove from $\mathcal{U}$ any object covered by $h^\star$\;
}
\Return sorted $\mathcal{Z}$\;
\end{algorithm}

\mypara{Instance-color assignment.}
Instance colors are used only to separate adjacent furniture instances of the same semantic class; they are not global instance ids. For each scene and semantic class $k$, we build an adjacency graph $G_k=(V_k,E_k)$ whose vertices are object instances of class $k$. An edge $(u,v)\in E_k$ is added when two different instances of class $k$ occupy horizontally adjacent grid cells in at least one height slice. \Cref{alg:instance_coloring} colors each $G_k$ independently, processing larger instances first and assigning each instance the first color not used by its already colored neighbors.

\begin{algorithm}[H]
\caption{Greedy instance-color assignment}
\label{alg:instance_coloring}
\KwIn{Semantic slices $\mathbf{Y}$; preprocessing instance-id slices $\mathbf{P}$; furniture classes $\mathcal{K}$; color palette $\Gamma=\{0,\ldots,C-1\}$}
\KwOut{Instance-color assignment $\gamma$ and conflict records $\mathcal{F}$}
$\gamma\leftarrow\emptyset$, $\mathcal{F}\leftarrow\emptyset$\;
\ForEach{semantic class $k\in\mathcal{K}$}{
  $V_k\leftarrow$ all instances that appear with class $k$\;
  $E_k\leftarrow\emptyset$\;
  \ForEach{height slice and horizontal neighbor pair $(p,q)$}{
    \If{$Y_p=Y_q=k$, $P_p\neq P_q$, and $P_p,P_q$ are valid furniture instance ids}{
      Add $(\min(P_p,P_q),\max(P_p,P_q))$ to $E_k$\;
    }
  }
  Sort $V_k$ by decreasing occupied cell count\;
  \ForEach{instance $u\in V_k$}{
    $N_u\leftarrow\{v:(u,v)\in E_k\text{ or }(v,u)\in E_k\}$\;
    $A\leftarrow\Gamma\setminus\{\gamma(v):v\in N_u\text{ and }v\text{ is colored}\}$\;
    \eIf{$A\neq\emptyset$}{
      $\gamma(u)\leftarrow$ the first color in $A$\;
    }{
      $\gamma(u)\leftarrow$ the first color in $\Gamma$\tcp*{fallback}
    }
  }
  $\mathcal{F}\leftarrow\mathcal{F}\cup\{(u,v)\in E_k:\gamma(u)=\gamma(v)\}$\;
}
\Return $\gamma,\mathcal{F}$\;
\end{algorithm}

The following single-slice result motivates using a small palette to distinguish adjacent instances instead of a separate label for every object.

\begin{theorem}[Four-color sufficiency for connected regions in a single slice]
\label{thm:instance_coloring}
Consider one height slice in which each object instance occupies a set of grid cells connected through shared edges. For each semantic class $k$, connect two instances when their cells share a grid edge. Then there exists an assignment of at most four colors such that every adjacent pair has different colors.
\end{theorem}

\mypara{Proof.}
The graph connecting cells that share an edge in a single slice is planar. Remove cells belonging to other semantic classes and contract each instance's connected cells into one vertex. Edge contraction and deletion preserve planarity, so the resulting instance-adjacency graph is planar. The four-color theorem therefore gives the required coloring~\citep{appel1977every,robertson1997four}.

We use this coloring principle to encode instance information with a compact palette. In preprocessing, greedy coloring assigns each instance one consistent color across slices. The combined adjacency graph across slices need not be planar, and greedy coloring is not guaranteed to find a four-color assignment. We choose three colors empirically: 99.08\% of adjacent same-category, different-instance token pairs have different colors (\cref{app:data_analysis}), and three colors yield lower COL and higher OC than four colors (\cref{tab:ablation_design}). The null label denotes an inapplicable attribute and is not a fourth instance color.

\section{Conditioning, Inference, and Scene Assembly Details}
\label{app:task_conditioning}
\label{app:inference}

\mypara{Arch-mask conditioning.}
The conditioning tuple in \cref{eq:multichannel_denoising} is written as $\kappa=(r,\mathbf{a})$ in the main text to keep the denoising notation compact. Here $r$ is the room type and $\mathbf{a}$ is the input arch mask. Cells with $a_i=1$ are clamped to their architectural labels during both forward masking and reverse denoising, while cells with $a_i=0$ are generated by the model. This lets \methodname{} sample semantic, instance-color, and facing fields for furniture while preserving the supplied room geometry.

\mypara{Sampling.}
 The three token fields are generated within one reverse diffusion process. Architectural tokens are fixed throughout; all unfixed cells initially have their semantic, instance-color, and facing fields masked. Each reverse step proceeds as follows:
\begin{enumerate}[leftmargin=*,itemsep=0pt,topsep=2pt]
\item Predict and update masked semantic labels according to the reverse schedule, retaining already revealed labels.
\item Recompute instance-color and facing predictions using the updated semantic state and shared denoiser features.
\item Keep both attributes masked if semantics remain masked. For revealed furniture semantics, reveal eligible attribute fields according to the schedule; for architecture/background semantics, assign null attributes.
\item Enforce compatible field values while preserving unresolved masks, and restore the fixed architectural tokens.
\end{enumerate}
For example, after a semantic label becomes ``chair,'' instance color or facing may be revealed in that step or remain masked until a later step. Such partially masked tokens are valid intermediate states. A mask means that a value is unknown; null means that an attribute is inapplicable. \Cref{app:token_validity} reports the denoising trajectory over 256 steps. The final grid preserves the supplied architectural tokens before grouping and asset placement.

\mypara{Token-to-object assembly.}
We group connected RoomTokens with the same semantic and instance-color labels across height slices into object instances. Each instance's tokens determine its category, horizontal center, and horizontal dimensions at grid resolution. The occupied height slices provide a coarse vertical extent, while the body/front labels indicate orientation.

MLP heads within \methodname{} estimate continuous height and vertical offset from its denoising features. For each instance, we form a 2D mask containing every grid cell where it appears in any height slice. We average the denoising features within this mask and feed the resulting vector to the heads. A yaw head provides orientation when the body/front labels do not determine the direction. These predictions, together with the attributes obtained directly from the tokens, specify each object's category, position, dimensions, and orientation for asset retrieval and placement.

\mypara{Scene assembly for field-removal ablations.}
With RoomTokens, our instance colors allow us to generate object instances with spatial extent and orientation information.  Our full model groups tokens by semantic and instance-color labels and uses the body/front cues to predict orientation.
To ablate the removal of fields such as instance-color and facing, we follow SemLayoutDiff~\cite{sun2026semlayoutdiff} on how to handle instance identification and orientation.
Without the instance color field, we use connected-component analysis to group connected tokens of the same semantic category into an object. In this case, adjacent same-category tokens are merged into the same instance.
Without the facing field, we use an auxiliary instance-level MLP head to predict the yaw from pooled instance features. Note that the yaw is not fixed to zero.
Asset retrieval and placement follow the same procedure for all of these variants.

\mypara{Asset retrieval.}
The final scene is instantiated by retrieving assets from 3D-FUTURE~\cite{fu20213dfuture}. For each object record, we first filter candidate assets by category, with a fallback to compatible coarse labels when exact labels are unavailable. We convert the predicted object size $(w,d,h)$ to 3D-FUTURE half-extents $(w/2,h/2,d/2)$ and select the candidate with the smallest squared error in this 3D size space. The selected asset is scaled  along each axis to match the estimated extents, rotated by the predicted yaw, translated to the recovered object position, and exported together with the reconstructed floor, wall, door, and window elements. The category- and size-based retrieval and placement procedure follows SemLayoutDiff. We evaluate the resulting scenes directly, without subsequent clipping, optimization, or hand-designed correction. The reported zero OOB describes boundary compliance of these final scenes.

\clearpage
\mypara{Effects of rotation and scaling.}
Even at a fixed position, changing an object's yaw or size can make its bounding box extend past a wall (OOB), cause a collision (COL), or block a passage (NAV). In \cref{tab:assembly-placement}, \methodname{} placement gives zero OOB. With fixed yaw, 30.90\% of scenes contain an out-of-bounds object, COL rises from 12.75\% to 35.82\%, and NAV falls from 99.30\% to 90.92\%. In \cref{fig:assembly-placement}(a), the rotated furniture collides with the coffee table and cuts off part of the walkable floor. Without scaling, scene OOB and COL rise to 10.50\% and 19.85\%, respectively, and NAV falls to 97.02\%. In \cref{fig:assembly-placement}(b), the two TV stands overlap even though the floor in this example remains connected.

\input{tables/table-assembly-placement}
\input{figs/fig-assembly-placement}

\section{Additional Qualitative Results}
\label{app:additional_qualitative_results}

\input{figs/fig-qua-iso}
\input{figs/fig-qua-ours}

In \cref{fig:qualitative-isometric}, we provide examples of output scenes from DiffuScene~\citep{tang2024diffuscene}, MiDiffusion~\citep{hu2026mixed}, SemLayoutDiff~\cite{sun2026semlayoutdiff}, and \methodname{} using isometric projection.

In \cref{fig:qualitative-ours}, we provide more generated results by our method in isometric view projection.

%% file: tables/table-unconditional.tex
\begin{table}[H]
  \centering
  \caption{Room-type-conditioned unconditional generation on 3D-FRONT. Our two-stage pipeline first generates an architecture map which is passed as input condition to the frozen MDiRT layout generator.}
  \label{tab:unconditional}
  \resizebox{\textwidth}{!}{
  \begin{tabular}{l ccccc ccccc}
    \toprule
    & \multicolumn{5}{c}{Distribution / Composition} & \multicolumn{5}{c}{Spatial / Physical Plausibility} \\
    \cmidrule(lr){2-6} \cmidrule(lr){7-11}
    Method & \fid & \kid & \ckl & Pair-Rel$\downarrow$ & \#Obj. & \oobscene & \oobobj & \col & \nav & OC$\uparrow$ \\
    \midrule
    \semdiff{} (unconditional) & 93.93 & \textbf{10.72} & \textbf{17.21} & 29.80 & 7.34 & 0.20 & 0.02 & 14.32 & 96.87 & 57.40 \\
    \textbf{ArchGen $\rightarrow$ MDiRT (ours)} & \textbf{69.96} & 12.86 & 18.35 & \textbf{21.70} & 8.27 & \textbf{0.00} & \textbf{0.00} & \textbf{12.59} & \textbf{99.55} & \textbf{81.10} \\
    \bottomrule
  \end{tabular}}
\end{table}

%% file: tables/table-per-room.tex
\begin{table*}[t]
  \centering
  \caption{Per-room-type evaluation on 3D-FRONT. Some living/dining source rooms occur in both category-specific breakdowns; the rows are therefore diagnostic subsets rather than a partition of the 390 distinct test rooms.}
  \label{tab:per_room}
  \resizebox{\textwidth}{!}{
  \begin{tabular}{ll rrrrr rrrrr}
    \toprule
    & & \multicolumn{5}{c}{Distribution / Composition} & \multicolumn{5}{c}{Spatial / Physical Plausibility} \\
    \cmidrule(lr){3-7} \cmidrule(lr){8-12}
    Room type & Method & \fid & \kid & \ckl & Pair-Rel$\downarrow$ & \#Obj. & \oobscene & \oobobj & \col & \nav & OC$\uparrow$ \\
    \midrule
    Bedroom & \diffuscene{} & 103.16 & 47.46 & 40.89 & 13.54 & 7.80 & 55.10 & 20.77 & 37.69 & 91.98 & 56.80 \\
    & \midiff{} & 103.74 & 32.06 & 61.72 & 52.26 & 6.65 & 65.25 & 37.81 & 33.60 & 95.25 & 73.30 \\
    & \semdiff{} & 75.42 & \textbf{5.93} & \textbf{5.27} & 30.92 & 7.20 & 13.80 & 4.00 & 26.21 & 93.80 & 52.80 \\
    & \textbf{Ours} & \textbf{51.06} & 9.86 & 7.92 & \textbf{13.21} & 7.55 & \textbf{0.00} & \textbf{0.00} & \textbf{22.29} & \textbf{98.41} & \textbf{78.60} \\
    \midrule
    Dining room & \diffuscene{} & 87.45 & 21.14 & 8.92 & 44.15 & 8.35 & 71.50 & 12.79 & 42.65 & 96.07 & 60.80 \\
    & \midiff{} & 95.93 & 36.47 & 43.71 & 47.79 & 7.10 & 46.63 & 30.74 & 64.62 & 97.22 & 77.10 \\
    & \semdiff{} & 70.87 & 9.97 & \textbf{4.97} & 16.80 & 7.75 & 15.80 & 7.53 & \textbf{5.37} & 99.36 & 58.20 \\
    & \textbf{Ours} & \textbf{49.78} & \textbf{7.81} & 5.38 & \textbf{10.47} & 8.05 & \textbf{0.00} & \textbf{0.00} & 11.22 & \textbf{99.59} & \textbf{83.50} \\
    \midrule
    Living room & \diffuscene{} & 74.79 & 22.06 & 11.47 & 39.78 & 9.80 & 72.80 & 26.53 & 42.57 & 94.98 & 60.78 \\
    & \midiff{} & 80.87 & 24.66 & 5.48 & 21.90 & 8.30 & 70.15 & 35.44 & 56.64 & 96.10 & 77.45 \\
    & \semdiff{} & 66.90 & \textbf{10.05} & \textbf{4.10} & 14.65 & 8.72 & 20.30 & 8.32 & 16.82 & 95.92 & 57.45 \\
    & \textbf{Ours} & \textbf{48.68} & 13.56 & 5.99 & \textbf{12.71} & 9.48 & \textbf{0.00} & \textbf{0.00} & \textbf{4.73} & \textbf{99.90} & \textbf{84.02} \\
    \bottomrule
  \end{tabular}}
\end{table*}

%% file: tables/table-novelty.tex
\begin{table}[t]
  \centering
  \caption{Conditional diversity and nearest-training analysis for 300 conditions with 10 seeds each (3,000 generations).}
  {
  \begin{tabular}{lc}
    \toprule
    Metric & Result \\
    \midrule
    Furniture $1-\mathrm{IoU}$ across seeds & 0.3227 \\
    Exact same-condition duplicate pairs & 0.5111\% \\
    Disagreement with paired ground truth & 0.6689 \\
    Generated $\rightarrow$ training median distance & 5.229 \\
    Held-out GT $\rightarrow$ training median distance & 4.938 \\
    Exact generated training duplicates & 1.3333\% \\
    Exact held-out GT training duplicates & 4.0000\% \\
    \bottomrule
  \end{tabular}}
  \label{tab:novelty}
\end{table}

%% file: tables/table-user-study-per-room.tex
\begin{table}[t]
  \centering
  \caption{User-study results by room type. Lower mean rank is better; the best mean rank in each room type is bold.}
  \label{tab:user_study_per_room}
  \resizebox{\textwidth}{!}{
  \begin{tabular}{lrrrrr}
    \toprule
    & \multicolumn{4}{c}{Mean rank$\downarrow$} & \methodname{} \\
    Room type & \methodname{} & \semdiff{} & \diffuscene{} & \midiff{} & First-place (\%)$\uparrow$ \\
    \midrule
Bedroom & \textbf{1.24} & 2.16 & 3.19 & 3.41 & 81.7 \\
Living room & \textbf{1.42} & 1.92 & 3.17 & 3.50 & 66.7 \\
Dining room & \textbf{1.20} & 2.23 & 3.08 & 3.49 & 82.1 \\
    \bottomrule
  \end{tabular}
  }
\end{table}

%% file: tables/table-representation-stats.tex
\begin{table}[h]
  \centering
  \caption{ Data analysis for the processed 3D-FRONT RoomToken representation.
  }
  \label{tab:representation_stats_app}
  \resizebox{\textwidth}{!}{
  \begin{tabular}{l r ccccccc}
    \toprule
    Data & Scenes & Object coverage & Invalid token & Color conflict & Inst. count MAE & Inst. exact & B/F paired & Orient MAE \\
    \midrule
    3D-FRONT (all splits) & 5{,}160 & 99.47\% & 0.00\% & 0.92\% & 0.18 & 91.53\% & 94.88\% & 1.13$^{\circ}$ \\
    \bottomrule
  \end{tabular}
  }
\end{table}

%% file: tables/table-assembly-placement.tex
\begin{table}[htbp]
  \centering
  \caption{Effect of rotation and scaling on the same generated scenes. All rows use the same retrieved assets at the same positions. Fixed yaw sets every object's yaw to zero. No scaling uses each asset at its retrieved size. An object is out of bounds if its placed bounding box extends outside the room. KID and CKL are scaled by $10^3$ and $10^2$.}
  \label{tab:assembly-placement}
  \setlength{\tabcolsep}{3pt}
  \resizebox{\linewidth}{!}{%
  \begin{tabular}{lrrrrrrrrrr}
    \toprule
    Asset placement & \fid & \kid & \ckl & Pair-Rel$\downarrow$ & \#Obj. & \oobscene & \oobobj & \col & \nav & OC$\uparrow$ \\
    \midrule
    \methodname{} placement & \textbf{49.84} & \textbf{10.41} & \textbf{6.43} & \textbf{12.13} & 8.36 & \textbf{0.00} & \textbf{0.00} & \textbf{12.75} & \textbf{99.30} & \textbf{82.04} \\
    Fixed yaw & 65.56 & 17.45 & \textbf{6.43} & 18.09 & 8.36 & 30.90 & 20.64 & 35.82 & 90.92 & 79.42 \\
    No scaling & 58.20 & 14.43 & \textbf{6.43} & 12.34 & 8.36 & 10.50 & 5.38 & 19.85 & 97.02 & 80.17 \\
    \bottomrule
  \end{tabular}}
\end{table}

%% file: figs/fig-assembly-placement.tex
\begin{figure}[htbp]
  \centering
  \includegraphics[width=\columnwidth]{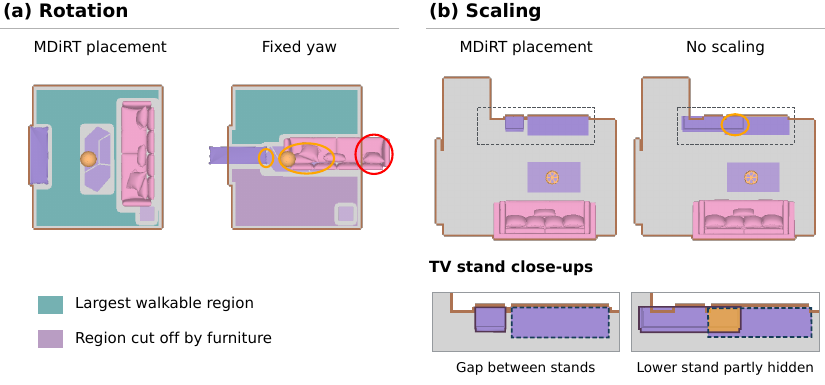}
  \caption{Effect of rotation and scaling on asset placement. Both original \methodname{} scenes have collision-free furniture placements and no measured boundary violations. Orange circles mark object--object intersections and red marks an object--wall intersection, as in \cref{fig:qualitative}. (a) With fixed yaw, the sofa and TV stand collide with the coffee table and split the walkable floor, lowering NAV. The floor overlay shows the largest connected walkable region in teal and the region cut off by the furniture in purple. (b) A gap separates the TV stands with \methodname{} placement. Without scaling, the taller stand overlaps the lower one. The dashed outline traces the partly hidden stand, and orange shading marks the overlap. This collision raises COL while the walkable floor remains connected.}
  \label{fig:assembly-placement}
\end{figure}

%% file: figs/fig-qua-iso.tex
\begin{figure}[t]
  \centering
  \includegraphics[width=\columnwidth]{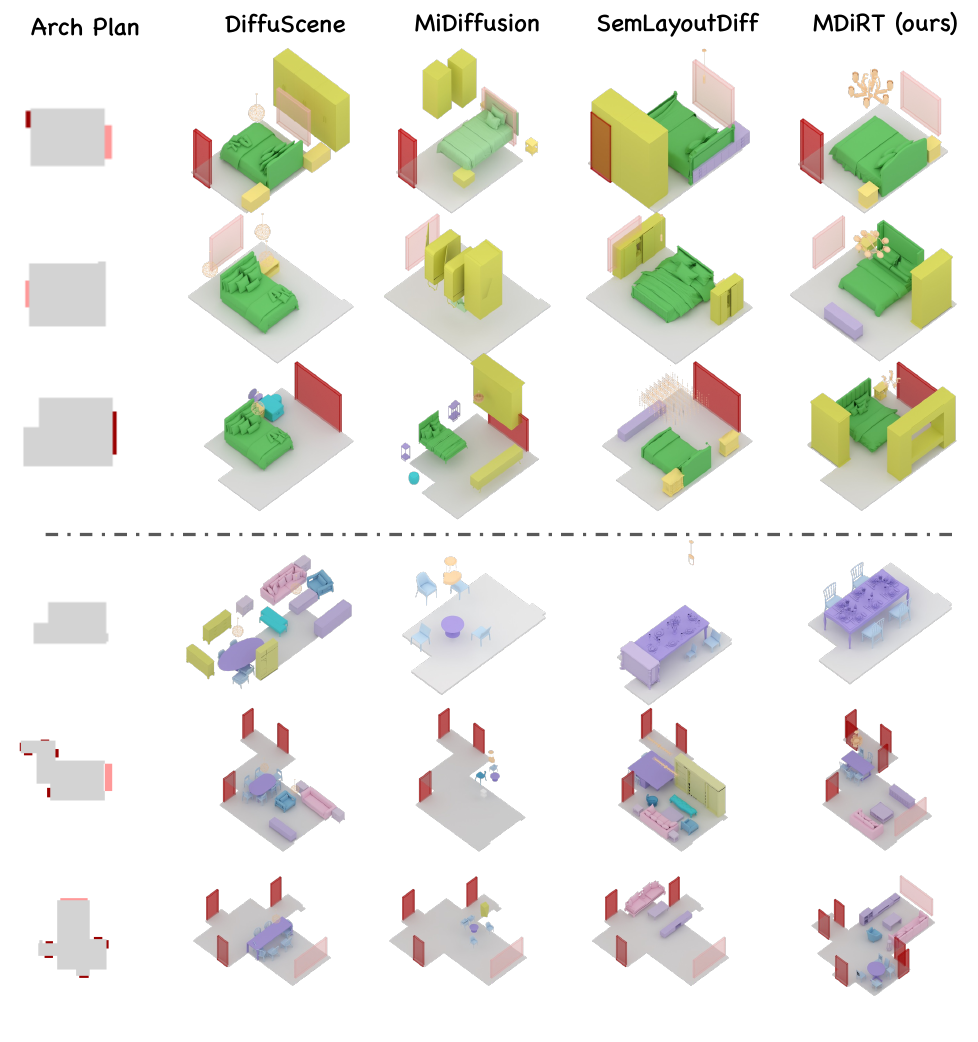}
  \caption{Arch-mask-conditioned qualitative comparison on 3D-FRONT in isometric view.}
  \label{fig:qualitative-isometric}
\end{figure}

%% file: figs/fig-qua-ours.tex
\begin{figure}[t]
  \centering
  \includegraphics[width=\columnwidth]{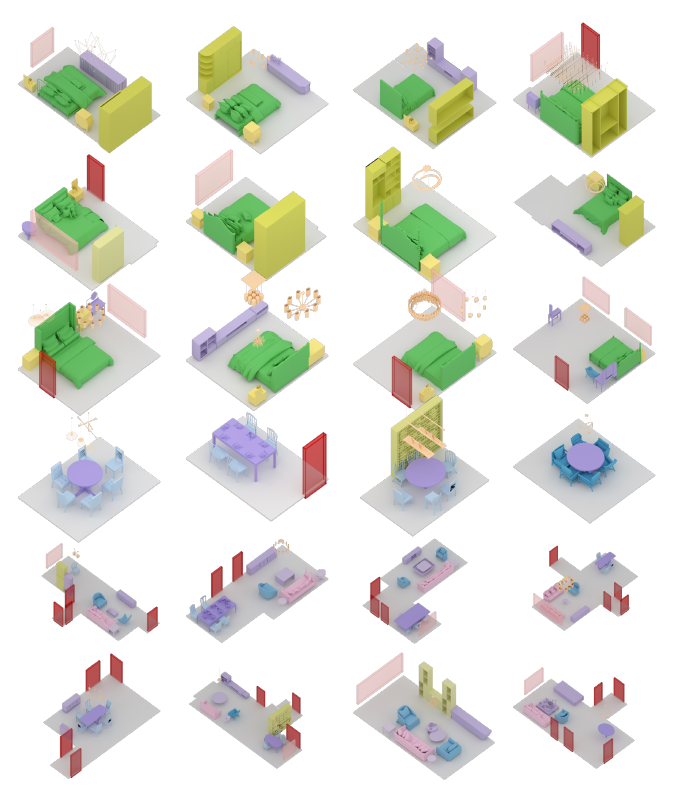}
  \caption{More arch-mask-conditioned generated results by \methodname{} in isometric view.}
  \label{fig:qualitative-ours}
\end{figure}